\documentclass{sig-alternate}
\usepackage{pbox}

\DeclareMathOperator*{\argmin}{arg\,min}

\begin{document}

\title{Improving Bilayer Product Quantization\\ for Billion-Scale Approximate Nearest Neighbors in High Dimensions}

\numberofauthors{2} 
%
\author{
%
%
\alignauthor
Artem Babenko \\
       \affaddr{Yandex}\\
       \affaddr{Moscow Institute of Physics and Technology}\\
       \email{artem.babenko@phystech.edu}
\alignauthor 
Victor Lempitsky\\
       \affaddr{Skolkovo Institute of Science and Technology}\\
       \email{lempitsky@skoltech.ru}
}

\newcommand{\fig}[1]{Figure~\ref{fig:#1}}
\newcommand{\sect}[1]{Section~\ref{sect:#1}}
\newcommand{\tab}[1]{Table~\ref{tab:#1}}
\newcommand{\eq}[1]{(\ref{eq:#1})}

\renewcommand{\dbltopfraction}{0.85}
\renewcommand{\textfraction}{0.10}
\renewcommand{\dblfloatpagefraction}{0.85}

\newcount\colveccount
\newcommand*\colvec[1]{
        \global\colveccount#1
        \begin{pmatrix}
        \colvecnext
}
\def\colvecnext#1{
        #1
        \global\advance\colveccount-1
        \ifnum\colveccount>0
                \\
                \expandafter\colvecnext
        \else
                \end{pmatrix}
        \fi
}

\maketitle
\begin{abstract}
The top-performing systems for billion-scale high-dimensional approximate nearest neighbor (ANN) search are all based on two-layer architectures that include an indexing structure and a compressed datapoints layer. An indexing structure is crucial as it allows to avoid exhaustive search, while the lossy data compression is needed to fit the dataset into RAM. Several of the most successful systems use product quantization (PQ)~\cite{Jegou11a} for both the indexing and the dataset compression layers. These systems are however limited in the way they exploit the interaction of product quantization processes that happen at different stages of these systems.

Here we introduce and evaluate two approximate nearest neighbor search systems that both exploit the synergy of product quantization processes in a more efficient way. The first system, called Fast Bilayer Product Quantization (FBPQ), speeds up the runtime of the baseline system (Multi-D-ADC) by several times, while achieving the same accuracy. The second system, Hierarchical Bilayer Product Quantization (HBPQ) provides a significantly better recall for the same runtime at a cost of small memory footprint increase. For the BIGANN dataset of billion SIFT descriptors, the 10\% increase in Recall@1 and the 17\% increase in Recall@10 is observed.
\\
\\
\end{abstract}




\section{Introduction}

The amount of data in multimedia applications grows rapidly and the methods for approximate nearest neighbor search (ANN) now routinely tackle billion-scale problems in high-dimensions. Several recent works \cite{Jegou11b,Babenko12,OpqTr} propose systems which are capable to handle huge datasets (such as the BIGANN dataset~\cite{Jegou11b} of a billion of 128-dimensional SIFT vectors) and perform searches in as little as few milliseconds in a single-thread mode, while using as little as eight gigabytes of memory.

All these systems consist of two layers: an indexing structure and compressed database points. An indexing structure allows to avoid exhaustive search while the lossy compression is required to fit high-dimensional points into RAM. 

In more details, the top-performing systems \cite{Jegou11b,Babenko12,OpqTr} split the search space into several \emph{cells}, which are essentially Voronoi cells for the \emph{codebook} of centroids. The system \cite{Jegou11b} forms this codebook via joint $K$-means clustering, while systems \cite{Babenko12,OpqTr} use product codebooks (based on product quantization~\cite{Jegou11a}) for cells centroid construction. For all points in each cell, their relative displacements from the cell centroid are then encoded, once again, via product quantization (PQ) \cite{Jegou11a}. 

At query time, the algorithm traverses cells which are close to a query. In each particular cell, the algorithm calculates Euclidean distances between the query displacement from the current cell centroid and the compressed displacements of database points stored in the cell. After visiting a predefined number of points, the  algorithm reranks them based on the distances calculated before. The systems \cite{Babenko12,OpqTr} thus use product quantization (PQ) \cite{Jegou11a} at both layers, although generally speaking the choice of indexing structure and the choice of compression method are independent and the PQ compression can be replaced by e.g.\ recent binary hashing method \cite{Zhang13,weiss08,Salakhutdinov09,Wang13}.

This work was inspired by the fact that the top-performing systems discussed above use the same global set of PQ codebooks for displacement compression in different index cells, while at the same time performing the search in each cell independently. This observation suggest two ways to boost the efficiency of search. The system can either (i) share some calculations between all cells and perform distance evaluation faster or, alternatively, (ii) use separate \emph{local} codebooks for each cell for better compression. Based on these considerations we propose and evaluate two systems: Fast Bilayer Product Quantization (FBPQ) and Hierarchical Bilayer Product Quantization (HBPQ).

FBPQ uses the fact that current systems \cite{Babenko12,OpqTr} do not use an advantage of fast asymmetric distance computation (ADC) with PQ \cite{Jegou11a}. In fact, they explicitly reconstruct displacements from their compressed representations so distance computation complexity becomes linear in data dimensionality. In this work, we show how to make distance computation complexity to be linear in the PQ code length reducing runtime substantially. We show experimentally that FBPQ performs several (up to $15$) times faster than the current state-of-the-art OMulti-D-OADC system \cite{OpqTr} providing the same recall levels.

HBPQ allows to achieve substantially higher recall comparing with OMulti-D-OADC (the previous to performer) and does this at a very slight increase in the computational cost and memory. We show below that distributions of displacements in different cells are quite different and global codebooks of \cite{Babenko12,OpqTr} provide suboptimal compression quality. This leads to idea that it would be gainful to have separate \emph{local} codebooks for each cell as they can adapt to particular displacements distribution. At a first glance, this might require too much memory (linear in number of cells if implemented naively) to store local codebooks. However, here we describe how the usage of PQ on both layers allows to construct and keep local codebooks with only modest increase in memory costs.

Overall, FBPQ and HBPQ provide a new state-of-the-art performance (both in terms of runtime and recall) of approximate nearest neighbor search for very large datasets of high-dimensional vectors such as the well known BIGANN dataset~\cite{Jegou11b}.

\section{Related work}

\begin{table*}
\centering
\renewcommand\arraystretch{1.3}
\begin{tabular}{|c|c|c|c|}
\hline
{\bf System} & {\bf Multi-D-ADC} & {\bf FBPQ} & {\bf HBPQ} \\
\hline
Indexing structure & inverted multi-index & inverted multi-index & inverted multi-index \\
\hline
Codebooks for database encoding & global & global & local \\
\hline
Distance evaluation complexity & $O(D)$ & $O(M)$ & $O(D)$ \\
\hline
Query preprocessing complexity & $O(\sqrt{C}D)$ & $O(\sqrt{C}D+KD)$ & $O(\sqrt{C}D)$ \\
\hline
\end{tabular}
\label{tab:systems_comparison}
\caption{The main differences between the current state-of-the-art Multi-D-ADC system, Fast Bilayer Product Quantization (FBPQ) and Hierarchical Bilayer Product Quantization (HBPQ). Here, $C$ is the number of the index cells (hundreds of thousands to hundreds of millions), $D$ is the dimensionality of the space (hundreds), $M$ is the number of the product quantization components (e.g.\ $8$), $K$ is the size of codebook for each PQ component (typically set to $256$). The strong advantage of the FBPQ is a small complexity of distance evaluation at a cost of cheap query preprocessing. The HBPQ has the same runtime as the Multi-D-ADC but uses local PQ codebooks for database compression, and hence provides better accuracy.}
\end{table*}

In this section we briefly cover several ideas from the previous work that are essential to the description of the proposed systems. Along the way, we introduce notation for this description.

\subsection{Product quantization}
Product quantization (PQ) is a lossy compression scheme for high-dimensional vectors~\cite{Jegou11a}. PQ encodes each vector $x \in \mathbf{R}^D$ as a concatenation of $M$ codewords from $M$ $\frac{D}{M}$\nobreakdash-dimensional codebooks $C_1,\ldots,C_M$, each containing $T$ codewords. In other words, PQ decomposes a vector into $M$ separate subvectors and applies vector quantization (VQ) to each subvector, while using a separate codebook. As a result each vector $x$ is encoded by a tuple of codewords indices $[i_1,\ldots,i_M]$ and approximated by $x \approx [C_1(i_1),\ldots, C_M(i_M)]$.
Fast Euclidean distance computation becomes possible via efficient \emph{ADC procedure} \cite{Jegou11a} using lookup tables:
\begin{gather}
\label{eq:adc}
\|q - x\|^2 \approx \|q - [C_1(i_1),\ldots, C_M(i_M)]\|^2 = \\
\sum\limits_{m=1}^{M}{\|q_m - C_m(i_m)\|}^2 \nonumber
\end{gather}
where $q_m$ --- $m$th subvector of a query $q$. This sum can be calculated in $M$ additions and lookups given that distances from query subvectors to codewords are precomputed. Thus, the ADC process requires $O(M \cdot T \cdot \frac{D}{M}) = O(TD)$ operations for lookup tables precomputations and $O(M)$ operations for distance evaluation. For compression purposes, $T$ is usually taken small, with $T=256$ being a popular choice (to fit index value into one byte). The precomputation time is thus negligible compared to the second stage, when distances to a large number of points are calculated.

From the geometry viewpoint, PQ effectively partitions the original vector space into $T^M$ cells, each being a Cartesian product of $M$ lower-dimensional cells. Such product-based approximation works better if the $\frac{D}{M}$-dimensional components of vectors have independent distributions. The degree of dependence is affected by the choice of the splitting, and can be further improved by orthogonal transformation applied to vectors as preprocessing. For some kinds of data partitions helps, and two recent works have looked into finding an optimal transformation \cite{Opq13,Norouzi13}. The modification corresponding to such pre-processing transformation is referred below as \emph{Optimized Product Quantization} (OPQ).

As a side note, it was shown in \cite{Jegou11a,Norouzi13} that the accuracy of the PQ compression considerably outperforms binary hashing methods for the same compression rates.

\subsection{The IVFADC system}
The first system capable of dealing with billion-scale datasets efficiently was {\emph IVFADC} introduced in \cite{Jegou11b}. This system combines the inverted index at the first (indexing) layer and product quantization at the second (compressed dataset) layer. IVFADC first splits the space into $C$ cells via the standard $K$-means and then encodes displacements of each point from the centroid of a cell it belongs to. The encoding is performed via product quantization that uses global codebooks shared by all cells. The authors of \cite{Jegou11b} mentioned that it would be possible to use different (i.e.\ local) codebooks for different cells but then memory consumption would be too large.
With IVFADC, the number of local codebooks would be linear in number of cells hence for large-scale problems their usage will lead to a dramatically increased memory consumption. Recent work \cite{Japan12} proposes to use multiple PQ codebooks each shared by several cells for the compression in IVFADC. \cite{Japan12} however provides experimental evaluation only for million-scale datasets and the scalability of this approach to billion-scale datasets is an open question.

\subsection{The inverted multi-index and Multi-D-ADC}
\label{sect:multidadc}
The inverted multi-index~\cite{Babenko12} is an indexing algorithm for high-dimensional spaces and very large datasets. The inverted multi-index generalizes the inverted index by using product codebooks for cells centroids construction (typically, as few as two components in the product are considered). Thus the inverted multi-index has two $\frac{D}{2}$-dimensional product codebooks for different halves of the vector, each with $T$ sub-codewords, thus effectively producing $C = T^2$ cells, where $C$ would typically be orders of magnitude bigger than the $C$ within the IVFADC system or other systems using inverted indices. Large number of cells provides very dense partitioning of the space, which means that a small fraction of dataset has to be traversed to achieve high recall (w.r.t.\ the true nearest neighbor). 

For dataset compression, \cite{Babenko12} followed the IVFADC system and used product quantization with global codebooks shared across all cells in order to encode the displacements of the vectors from centroids (this system is referred to as \emph{Multi-D-ADC}). 

For the sake of self-containment we give a brief description of the Multi-D-ADC system as next sections rely on it significantly. In particular, we focus on the second-order Multi-D-ADC but the generalization to other orders is straightforward.

We discuss three aspects/steps of the Multi-D-ADC: the learning of codebooks, the construction of the indexing structure, and the processing of queries.

\begin{table*}
\small
\centering
\addtolength{\tabcolsep}{-1pt}
\renewcommand\arraystretch{1.3}
\begin{tabular}{|c|c|ccc|c|c|c|}
\hline
System & $l$ &  R@1 & R@10 & R@100 & Time(ms) & Speed-up factor & Memory(Gb)\\
\hline
\multicolumn{8}{|c|}{\bf BIGANN, 1 billion SIFTs,  8 bytes per vector}\\
\hline
OMulti-D-OADC & 10000 & 0.179 & 0.523 & 0.757 & 4.9 & --- &13.0 \\
FBPQ & 10000 & 0.179 & 0.5234 & 0.757 & {\bf 1.9} & 2.6 & 13.15 \\
\hline
OMulti-D-OADC & 30000 & 0.184 & 0.549 & 0.853 & 13.8 & --- & 13.0\\
FBPQ & 30000 & 0.184 & 0.549 & 0.853 & {\bf 3.6} & 3.8 &13.15\\
\hline
OMulti-D-OADC & 100000 & 0.186 & 0.556 & 0.894 & 41.3 & --- & 13.0\\
FBPQ & 100000 & 0.186 & 0.556 & 0.894 & {\bf 9.7} & 4.3 & 13.15\\
\hline
\multicolumn{8}{|c|}{\bf GIST50M, 50 millions GISTs,  8 bytes per vector}\\
\hline
OMulti-D-OADC & 10000 & 0.317 & 0.454 & 0.569 &6.3 & --- & 0.57 \\
FBPQ & 10000 & 0.317 & 0.454 & 0.569 & {\bf 0.6} & 10.5 & 0.57 \\
\hline
OMulti-D-OADC & 30000 & 0.323 & 0.496 & 0.659 & 18.5 & --- & 0.57\\
FBPQ & 30000 & 0.323 & 0.496 & 0.659 & {\bf 1.4} & 13.2 & 0.57\\
\hline
OMulti-D-OADC & 100000 & 0.327 & 0.512 & 0.711 & 61.4 & --- & 0.57\\
FBPQ & 100000 & 0.327 & 0.512 & 0.711 & {\bf 4.2} & 14.6 & 0.57\\
\hline
\end{tabular}
\caption{Comparison of bilayer systems runtime: current state-of-the art OMulti-D-OADC and the optimized FBPQ on BIGANN and GIST50M datasets. $l$ is a number of candidates reranked by both systems. The optimized FBPQ provides the same recall levels as OMulti-D-OADC up to $4$ times faster on SIFT1B and up to $15$ times faster on GIST50M. Advantage on GIST50M is more impressive as GIST descriptors are $3$ times longer than SIFT.}
\label{tab:FBPQ}
\end{table*}
\addtolength{\tabcolsep}{-1pt}

\subsubsection{Multi-D-ADC codebooks learning}

We assume that a large set of $N$ $D$-dimensional learn vectors $L = \{p_1,\ldots,p_N\} \subset \mathbf{R}^D$ is given. Let $p_k = \left[p_k^1 \; p_k^2 \right]$ be the decomposition of an $k$-th training vector $p_k \in \mathbf{R}^D$ into two halves, $p_k^1 \in \mathbf{R}^{\frac{D}{2}}$, $p_k^2 \in \mathbf{R}^{\frac{D}{2}}$. Then, following the standard PQ practice, the codebook for the first half of dimensions  $C^1 = \{c_1^1, c_2^1, \ldots, c_T^1\}$ is obtained via the K-means clustering of a set $\{p_1^1,\ldots,p_N^1\}$. Analogously, the codebook for the second halves $C^2 = \{c_1^2, c_2^2, \ldots, c_T^2\}$ is obtained via the K-means clustering of a set $\{p_1^2,\ldots,p_N^2\}$.

Codebooks $C^1$ and $C^2$ define two $\frac{D}{2}$-dimensional \\ quantizers 
\begin{gather}
\label{eq:q1}
q^1 : \mathbf{R}^{\frac{D}{2}} \rightarrow C^1 \subset \mathbf{R}^{\frac{D}{2}} \\
q^1(x) = \argmin_{c_i^1 \in C^1}\|x - c_i^1\|^2 \nonumber
\end{gather}
and
\begin{gather}
\label{eq:q2}
q^2 : \mathbf{R}^{\frac{D}{2}} \rightarrow C^2 \subset \mathbf{R}^{\frac{D}{2}} \\
q^2(x) = \argmin_{c_i^2 \in C^2}\|x - c_i^2\|^2 \nonumber
\end{gather}

The quantizers $q^1, q^2$ define a partition of initial $D$-dimensional space into $T^2$ Voronoi cells $\{W_{ij}\}_{i,j=1}^T$:

\begin{gather}
\label{eq:Wij}
W_{ij} = \{ x = [x^1\;x^2]\in \mathbf{R}^{D} \mid \; q^1(x^1) = c_i^1, q^2(x^2) = c_j^1 \}
\end{gather}

For each point $p_k = [p_k^1\;p_k^2]$, a displacement from its cell centroid is calculated. Then, $M$ PQ codebooks $R_1,\dots,R_M$ for displacements compression are learned on the set of displacements from all cells:

\begin{gather}
\left\{ \colvec{2}{p_k^1}{p_k^2} - \colvec{2}{c_i^1}{c_j^2} \; \mid \;[p_k^1\;p_k^2] \in L,\;q^1(p_k^1) = c_i^1,\;q^2(p_k^2) = c_j^2 \right\}
\end{gather}

The size of codebooks $R_1,\dots,R_M$ is usually set to $256$ in order to fit each codeword id into one byte.

Overall, the set of Multi-D-ADC codebooks consists of

\begin{enumerate}
\item Two indexing-layer codebooks $C^1$ and $C^2$, \\ $|C^1| = |C^2| = T$
\item $M$ compression-layer codebooks $R_1,\dots,R_M$
\end{enumerate}

\subsubsection{Multi-D-ADC index construction}

Codebooks $C^1$ and $C^2$ subdivide the space into the cells $W_{ij}$ defined by product codebooks $C^1$ and $C^2$ as in \eq{Wij}.

All points from each cell $W_{ij}$ are stored contiguously in a one-dimensional array. All cells are also stored in a large array and each cell is represented by an integer keeping position of the starting position of the points belonging to a particular cell. This construction allows to retrieve all points from a few cells efficiently via multi-sequence algorithm \cite{Babenko12}. 

Each point in the Multi-D-ADC index is represented by the PQ-encoding of a displacement from the centroid of the cell it belongs to. Then for each point $x$ falling into the cell $W_{ij}$, Multi-D-ADC encodes the $D$-dimensional vector of displacement:
\begin{gather}
\label{eq:displacement}
d = x - \colvec{2}{c_i^1}{c_j^2} = \colvec{2}{x^1 - c_i^1}{x^2 - c_j^2}
\end{gather}

If the PQ encoding of $d$ with codebooks $R_1,\dots,R_M$ is $[r_1,\dots,r_M]$ then the initial point $x$ is effectively approximated by:

\begin{gather}
\label{eq:approximation}
x \approx \colvec{2}{c_i^1}{c_j^2} + \colvec{3}{r_1}{\vdots}{r_M}
\end{gather}

\subsubsection{ANN with Multi-D-ADC}

Once the Multi-D-ADC index is constructed, it can be used to perform fast approximate nearest neighbor search. Thus, given a query $q = \left[q^1\;q^2\right]\in \mathbf{R}^D $ the Multi-D-ADC index allows to get a ranked list of $l$ points which are likely to be neighbors of $q$. This list is created in two steps --- forming the set of $l$ candidates and reranking of candidates.

At first Multi-D-ADC identifies closest codewords for $q^1$ in $C^1$ and for $q^2$ in $C^2$. Then, the multi-sequence algorithm \cite{Babenko12} merges the two lists of closest codewords into ordered sequence of closest cells $\{W_{ij}\}$. The multi-sequence algorithm stops when the sufficient number of cells are traversed, i.e.\ they contain the required number of points $l$. These points form initial set of candidates.  

In the second step, the candidates are reranked based on their PQ codes. In each visited cell query displacement w.r.t.\ the cell centroid is calculated. Then distances from query displacement to database points displacements are evaluated. Multi-D-ADC do not use fast ADC procedure as it would require query displacement quantization in each cell which is too costly for a large number of cells. Instead, Multi-D-ADC reconstructs displacements explicitly and calculates each distance in $O(D)$ operations. Finally, the candidates are reranked according to the increasing distance from the query.

\subsubsection{OMulti-D-OADC}

\cite{OpqTr} further improved the performance of Multi-D-ADC by replacing product quantization with optimized product quantization for both indexing and compression (hence the name of their system \emph{OMulti-D-OADC}). OMulti-D-OADC gives the state-of-the-art performance in terms of the search accuracy on the BIGANN dataset (the only system that achieves higher accuracy \cite{Joint13} does this at the cost of considerable, e.g.\ five-fold increase in the overall memory usage, which is undesirable for most real-life scenarios).

\subsubsection{Bilayer Product Quantization}

To sum up, we notice that the Multi-D-ADC (OMulti-D-OADC) uses PQ on both indexing and compressed database layers, hence it is effectively a Bilayer Product Quantization system. At the same time PQ is not a necessary choice for compression in the Multi-D-ADC. E.g.\ one can employ any other compression scheme (e.g. binary embedding, etc.).

\section{Improving bilayer product quantization}

As described above, the Multi-D-ADC is a top-performing system for ANN search which allows to provide a short-list of candidates for a given query in a few milliseconds. In this section, we address the issue of efficient reranking of this short-list using the fact that system uses PQ on both layers. We introduce two extensions of the Multi\nobreakdash-D\nobreakdash-ADC: Fast Bilayer Product Quantization (FBPQ) and Hierarchical Bilayer Product Quantization (HBPQ) which allow to boost reranking efficiency either in terms of runtime or in terms of accuracy.

\begin{figure*}
\centering
\begin{tabular}{cccc}
\includegraphics[width=4.0cm]{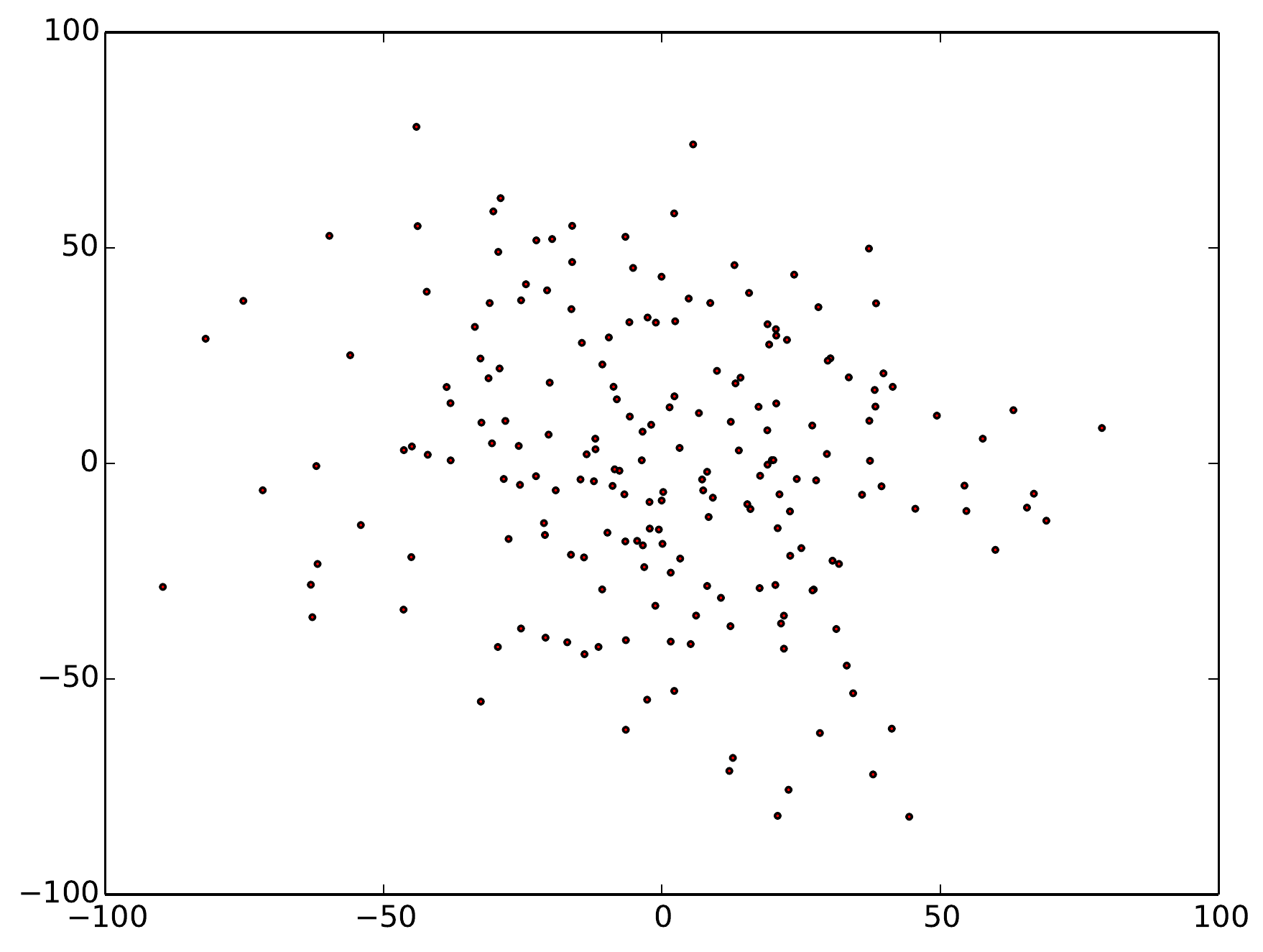}&
\includegraphics[width=4.0cm]{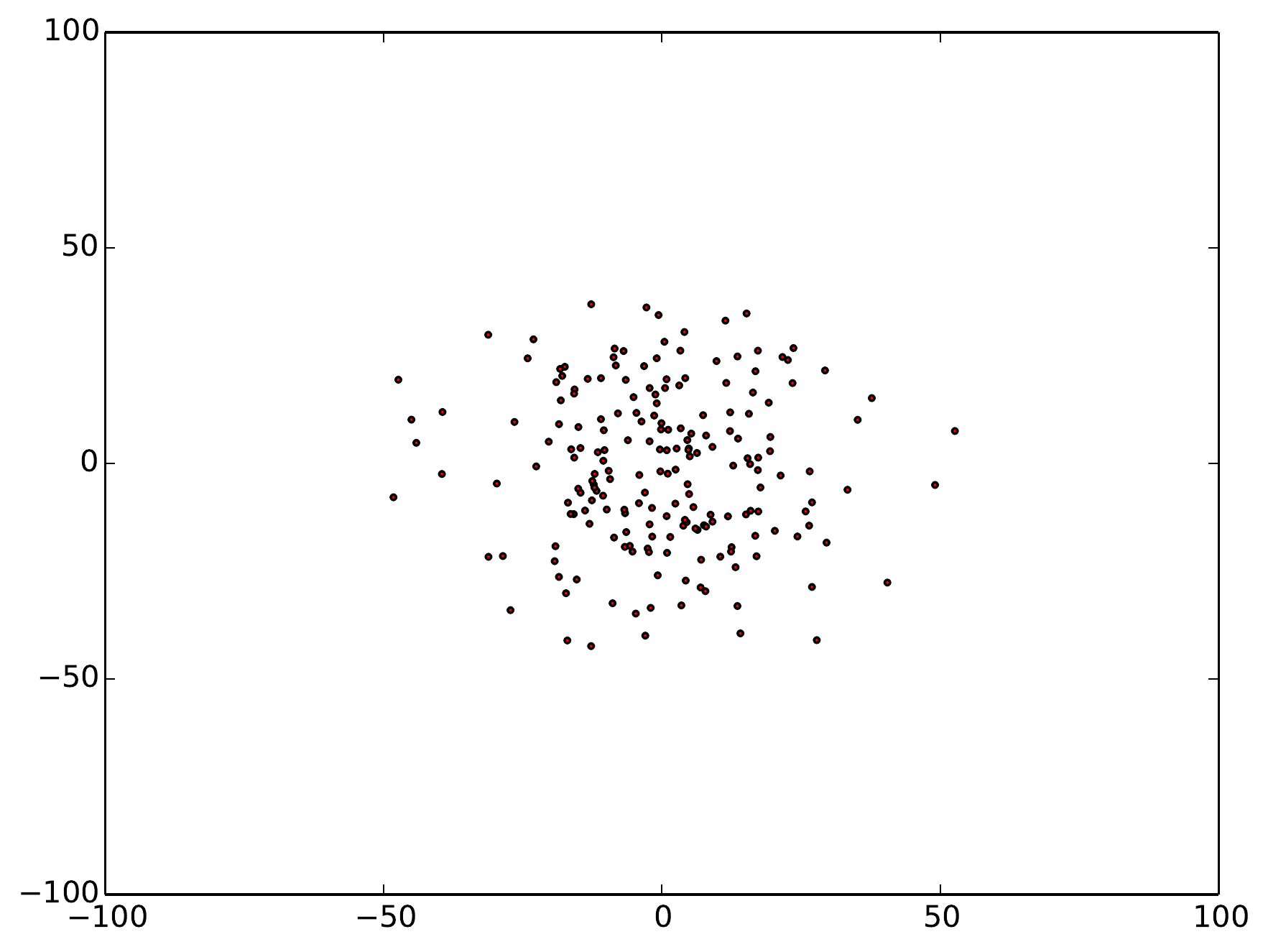}&
\includegraphics[width=4.0cm]{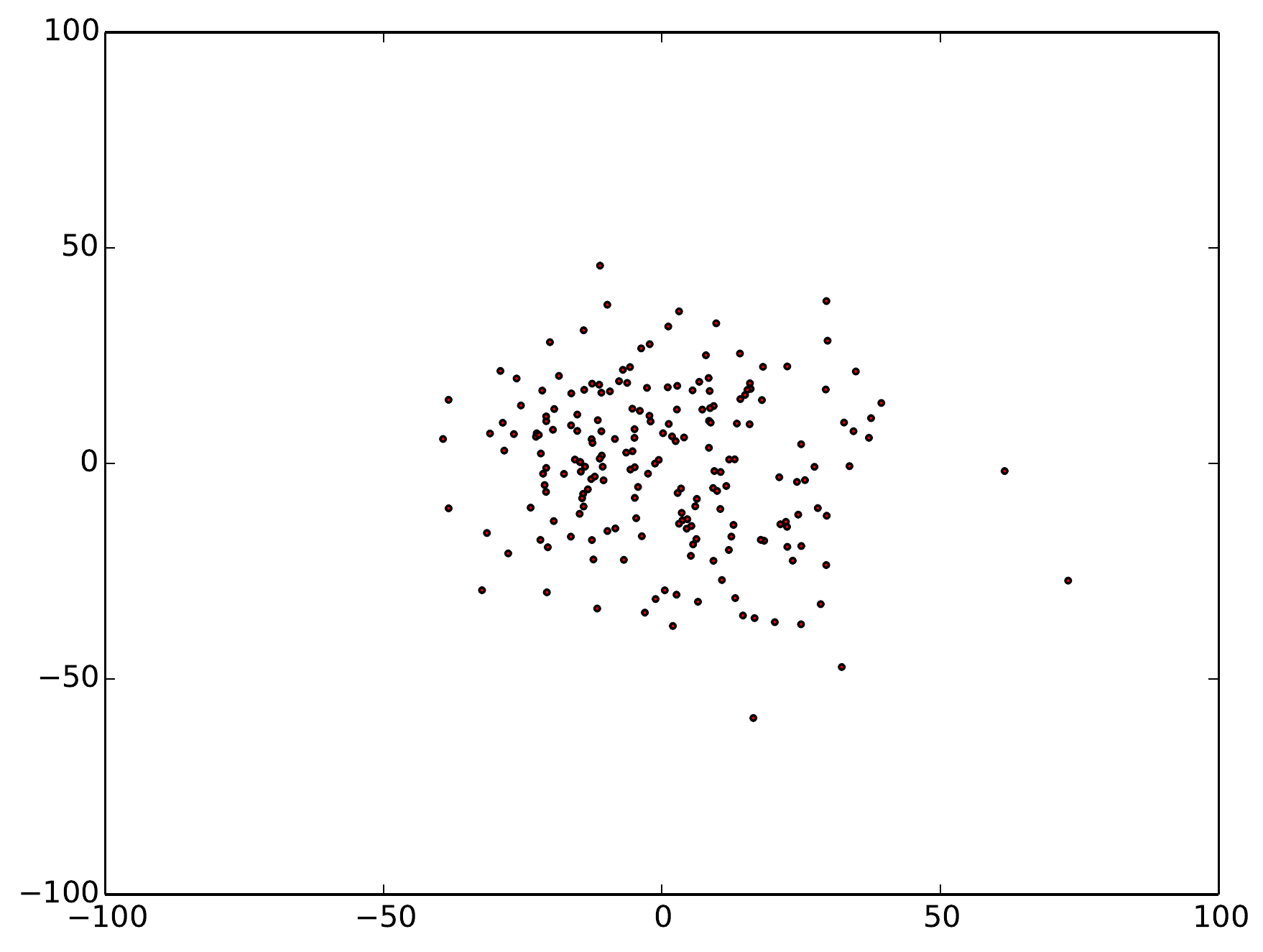}&
\includegraphics[width=4.0cm]{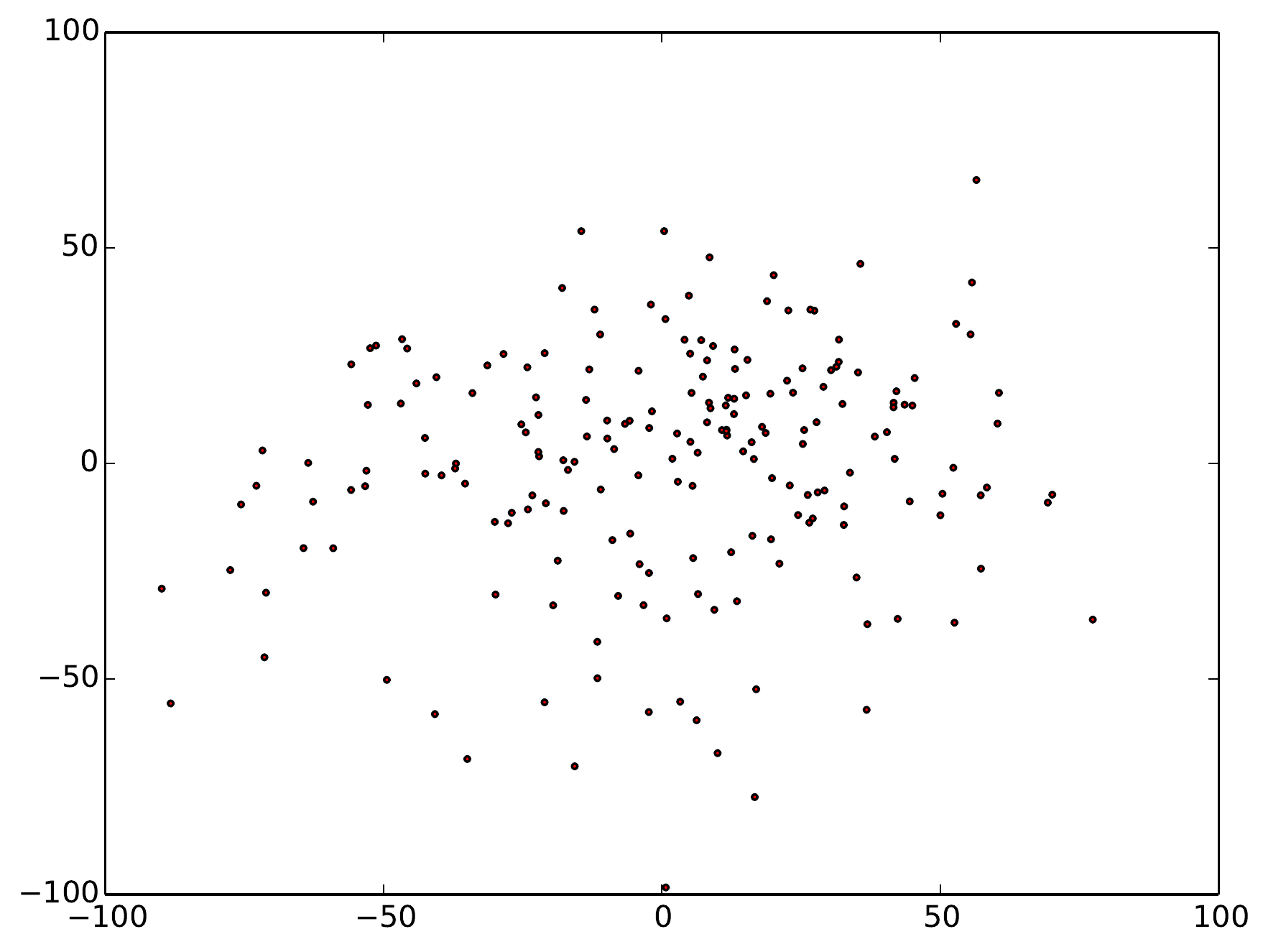}\\
\includegraphics[width=4.0cm]{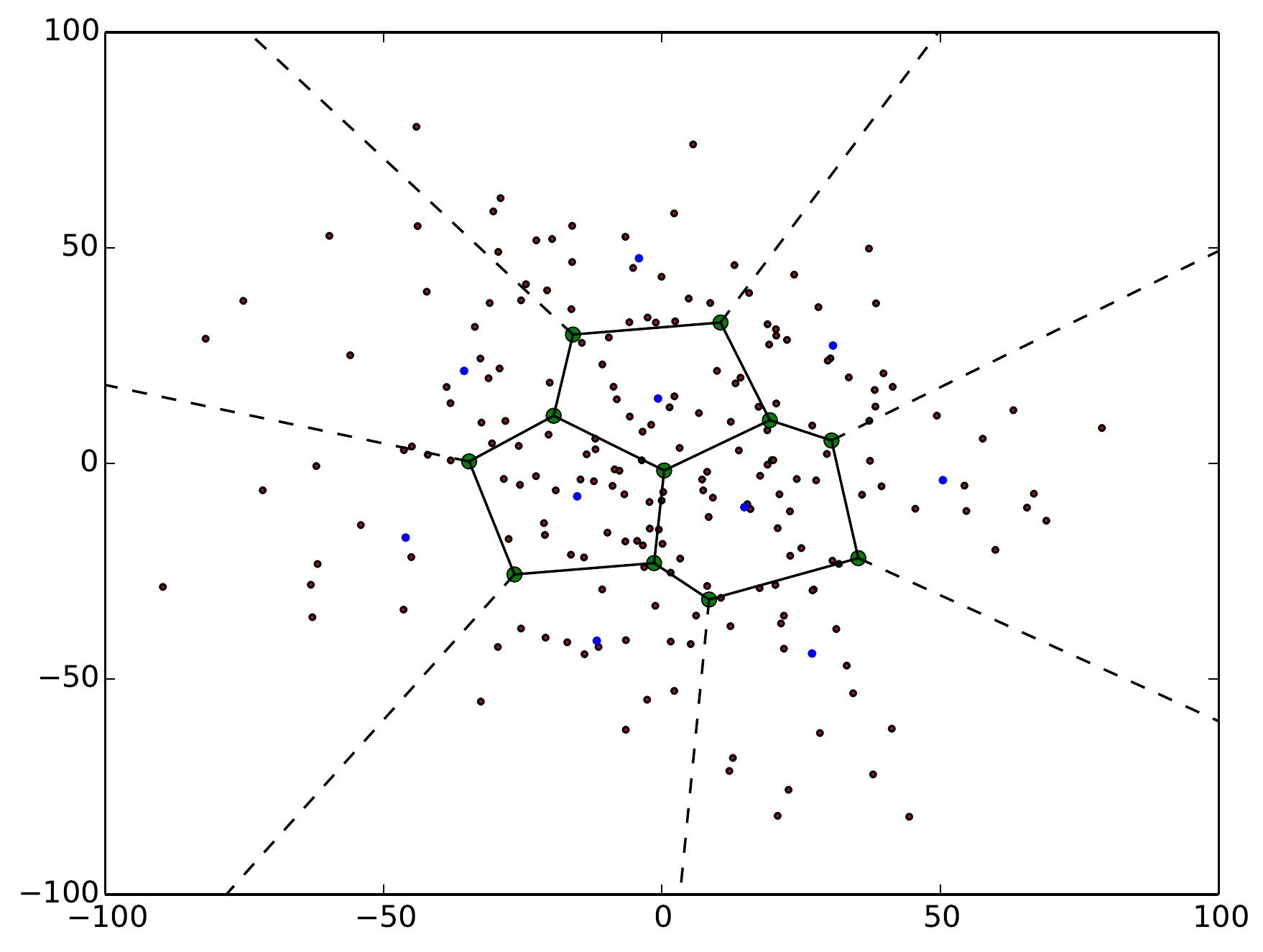}&
\includegraphics[width=4.0cm]{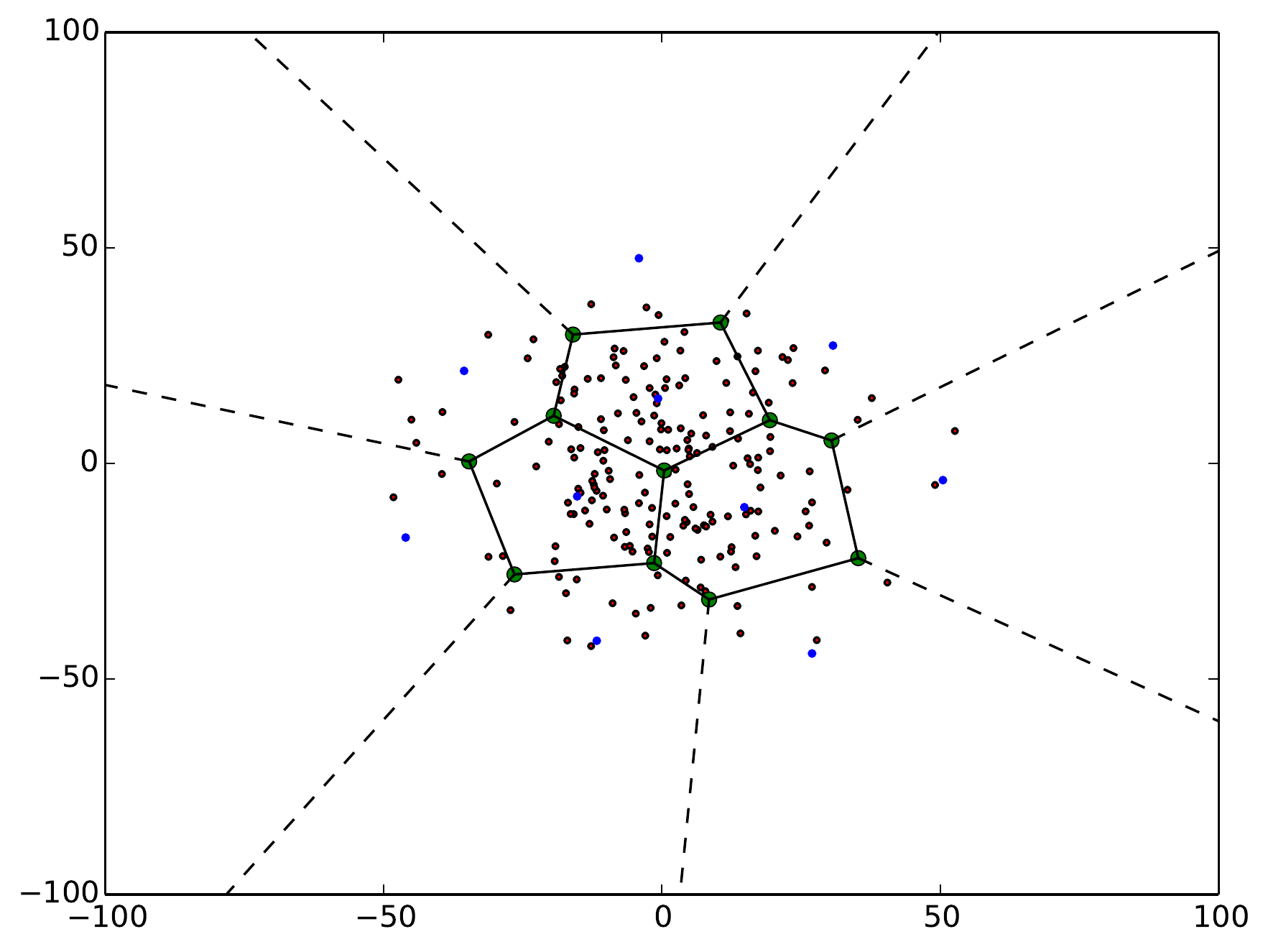}&
\includegraphics[width=4.0cm]{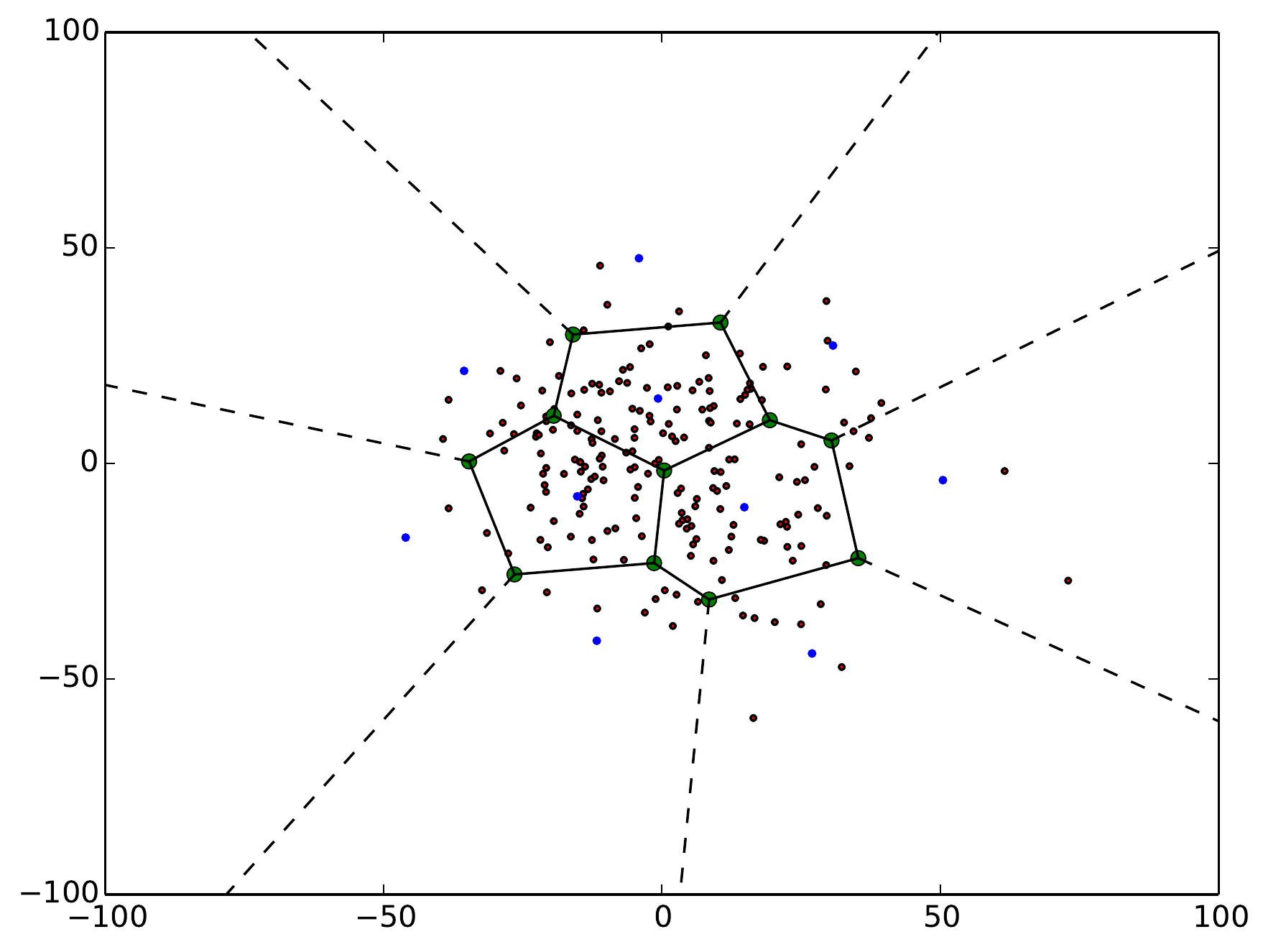}&
\includegraphics[width=4.0cm]{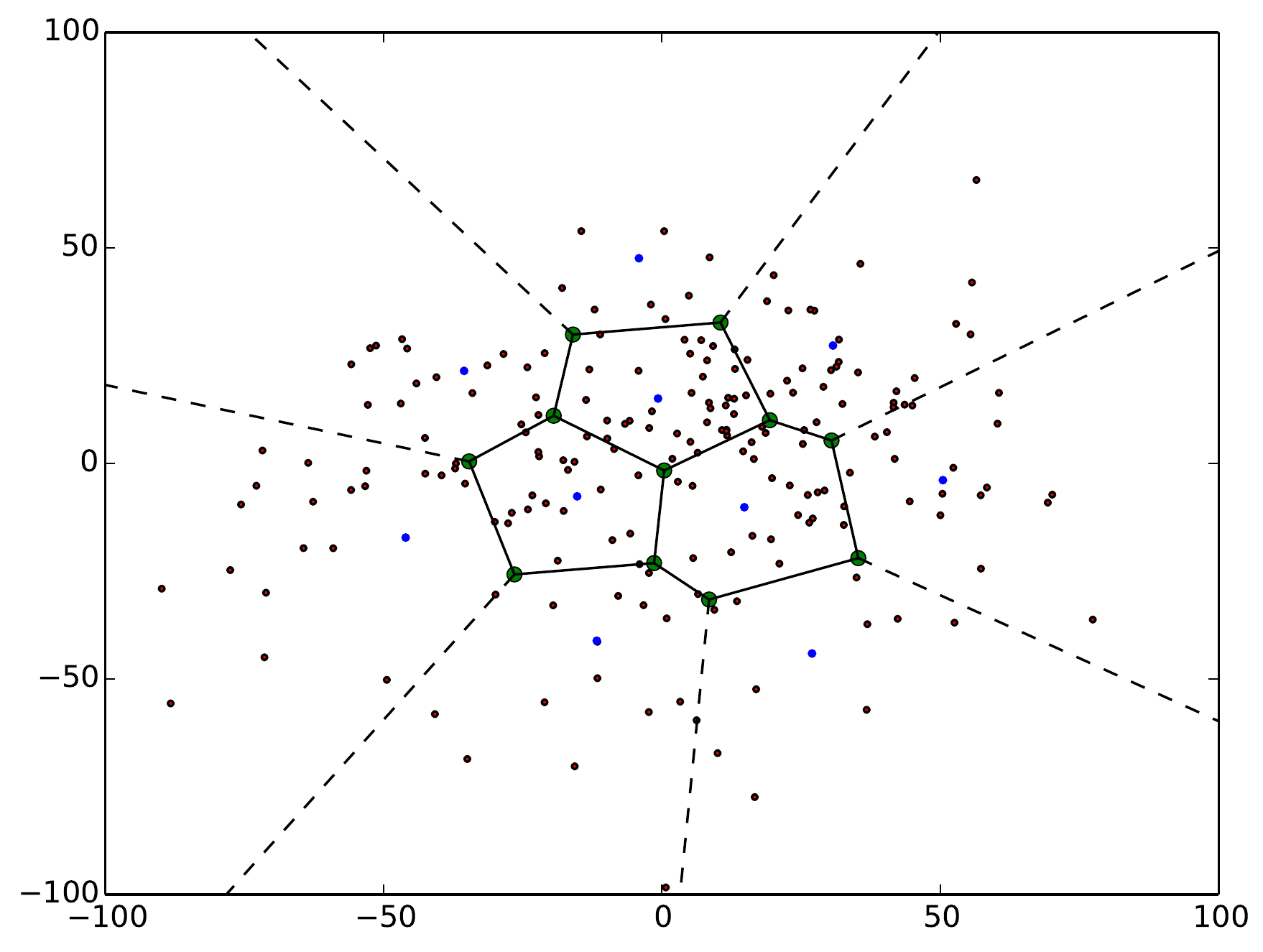}\\
\includegraphics[width=4.0cm]{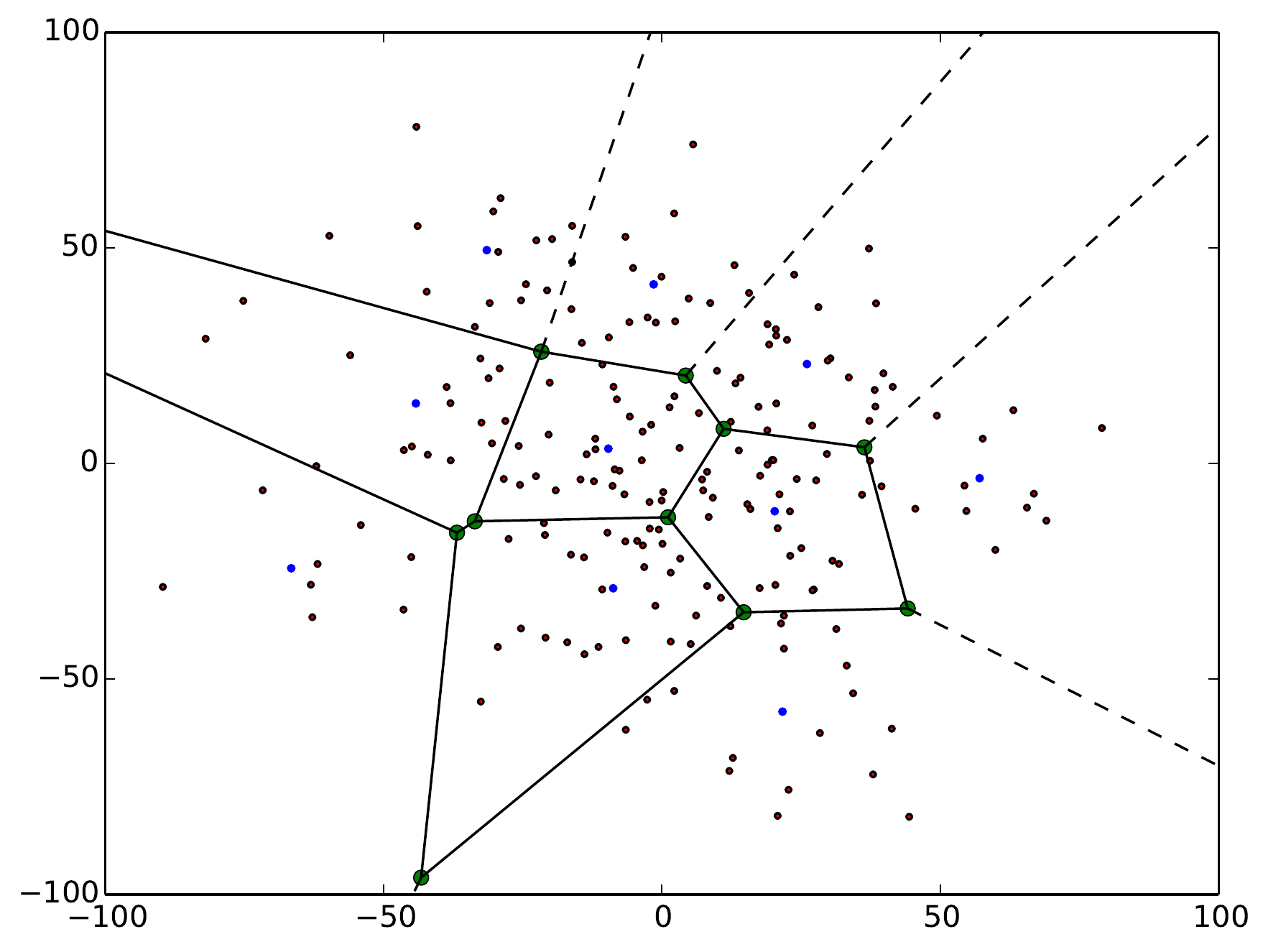}&
\includegraphics[width=4.0cm]{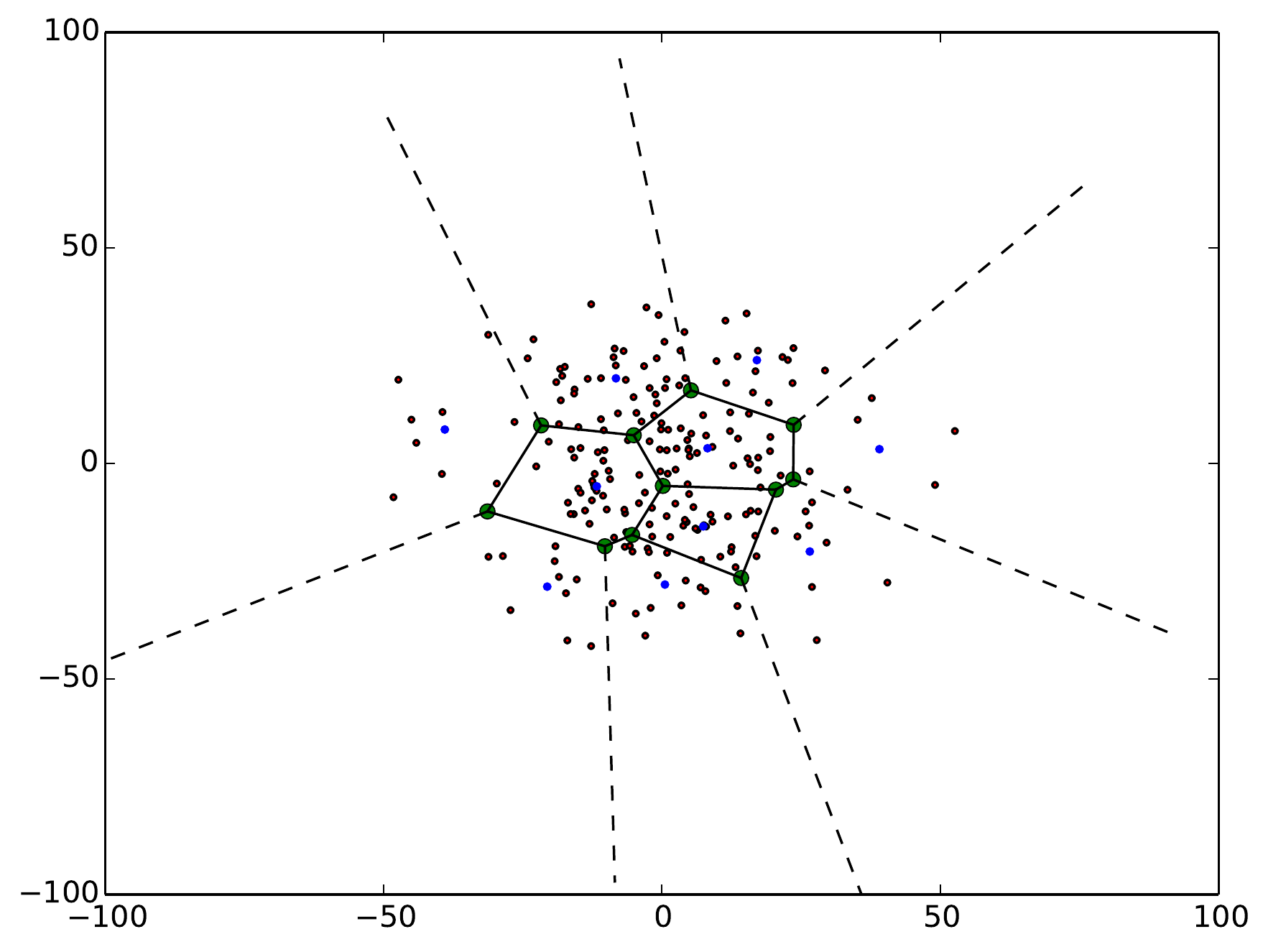}&
\includegraphics[width=4.0cm]{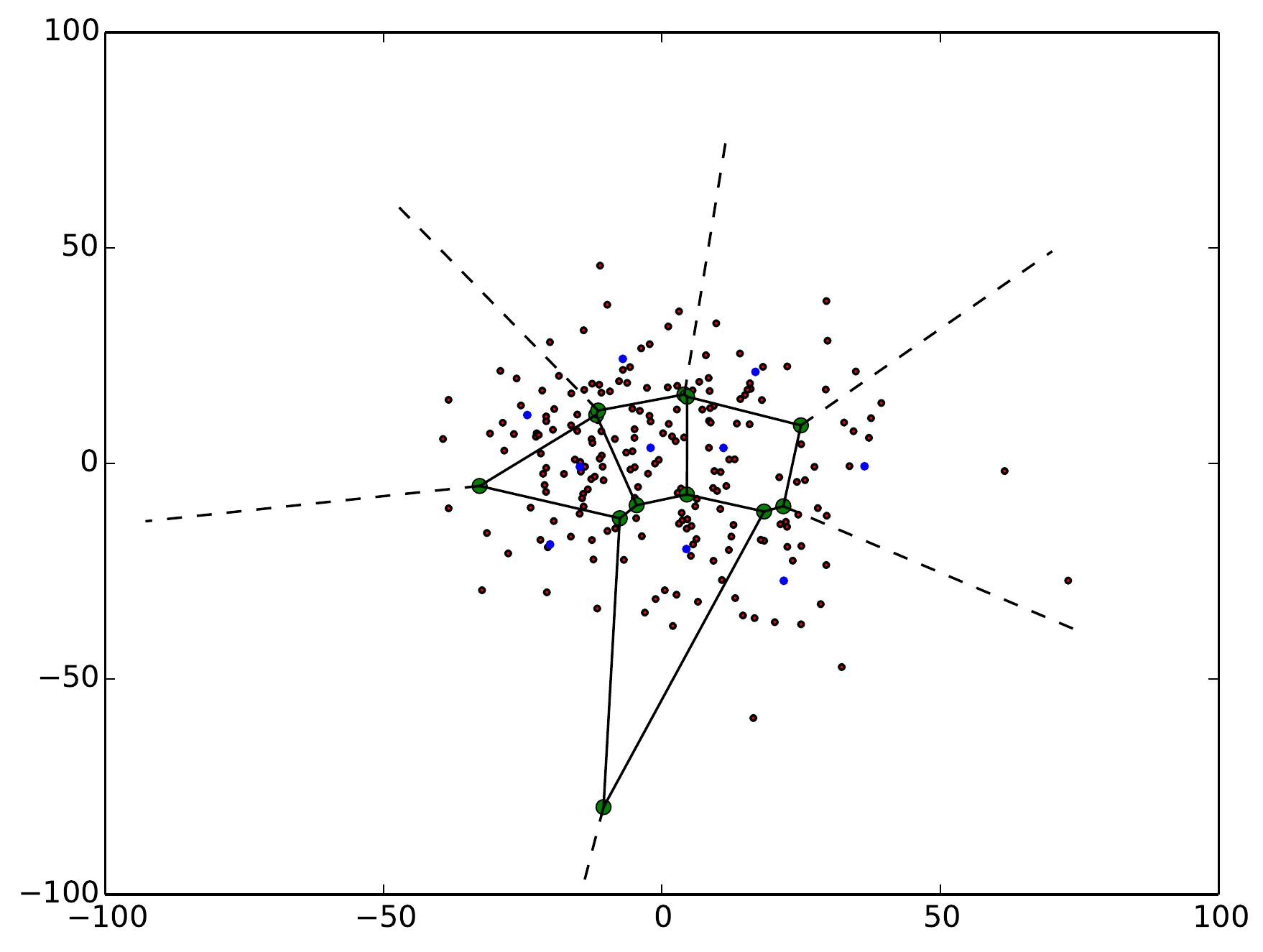}&
\includegraphics[width=4.0cm]{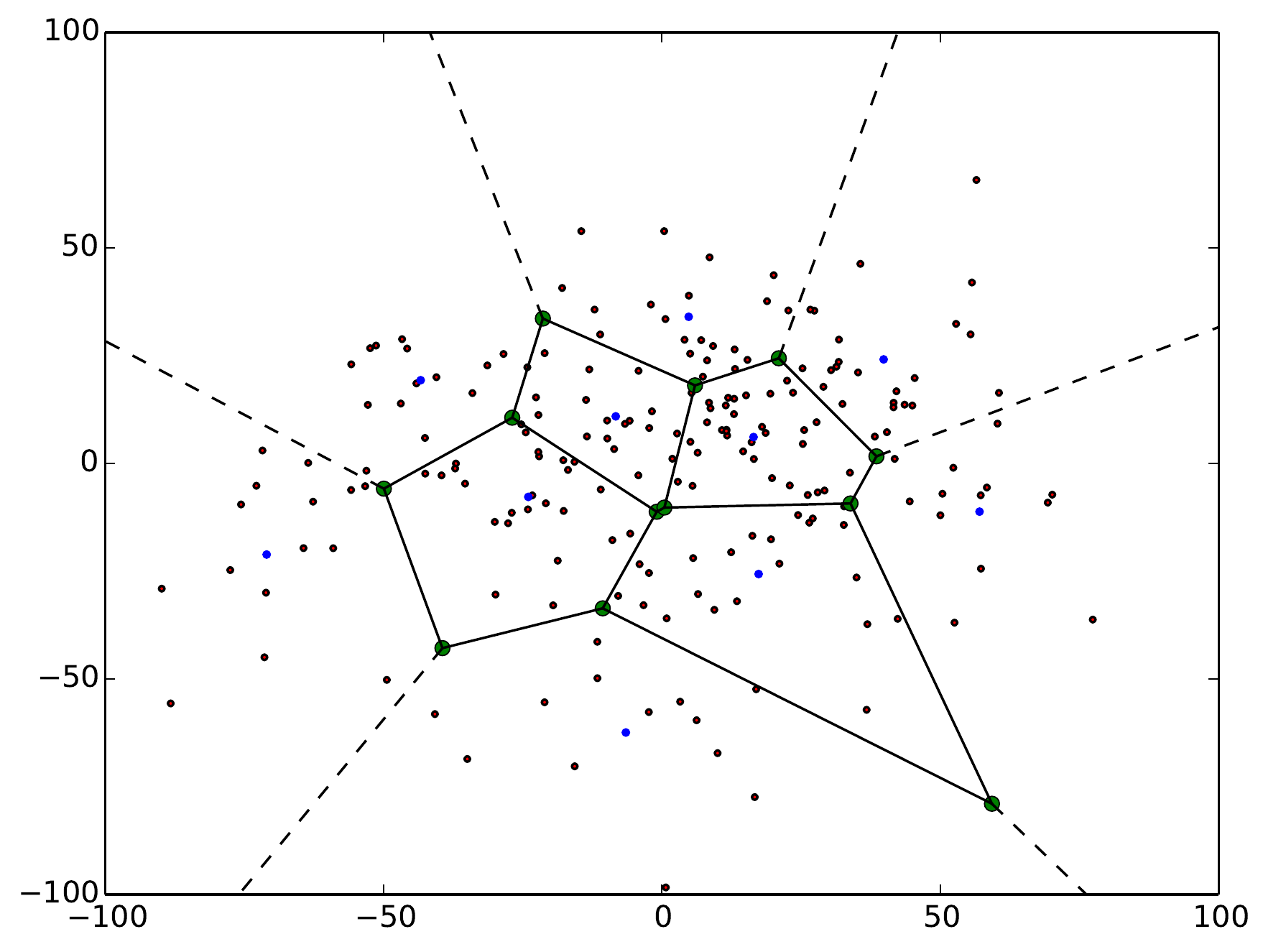}\\
\end{tabular}
\caption{Distributions of points displacements and Voronoi diagrams of global and local codebooks of size $10$ for $4$ different cells in the second-order inverted multi-index with $C = 2^{28}$ and the SIFT1B dataset. 200 random points from each cell and are visualized by their first two principal components. Codewords from both codebooks are indicated by blue points. First row demonstrates global codebooks which are learnt by $K$-means on the displacements from all cells. Second row corresponds to local codebooks which are learnt for each cell separately. There is a significant variation in the distributions and the usage of separate local codebooks in each cell can improve encoding substantially.}
\label{fig:global_vs_local}
\end{figure*}

\subsection{FBPQ}

The Multi-D-ADC system does not use the fact that PQ codebooks for displacements compression are shared by all cells. In other words, it does not share any calculations among cells and perform search in each cell independently. Hence evaluation of distance from a query to any database point requires explicit point displacement reconstruction and takes $O(D)$ operations.
It would be a natural improvement to use caching of distances in each cell which would allow to reuse distances to PQ subcodewords computed before. However, simple caching turns out to be inefficient in the case of the Multi-D-ADC as each cell contains only a few points and a number of cache hits is small.

We now describe a modification for the Multi-D-ADC which guarantees to speed up distance evaluation from $O(D)$ operations to $O(M)$ operations.

For this, let us consider Euclidean distance between query $q \in \mathbf{R}^D$ and a dataset point $x$ belonging to a cell $W_{ij}$ with centroid $[c^1_i, c^2_j]$. The displacement of $x$ from cell centroid is PQ-encoded as a concatenation $[r_1,\dots, r_M]$. Using the \eq{approximation} we get:

\begin{gather}
\label{eq:distance}
\|q - x\|^2 \approx \left\|q - \colvec{2}{c_i^1}{c_j^2} - \colvec{3}{r_1}{\vdots}{r_M}\right\|^2 = \\
||q||^2 - 2\left\langle q,\colvec{2}{c_i^1}{c_j^2} \right\rangle - 2\left\langle q, \colvec{3}{r_1}{\vdots}{r_M} \right\rangle + \left\|\colvec{2}{c_i^1}{c_j^2} + \colvec{3}{r_1}{\vdots}{r_M}\right\|^2 \nonumber
\end{gather}

The idea of the FBPQ is that dot-products of query subvectors and centroids from codebooks $C^1$, $C^2$ and $R_1,\dots,R_M$ can be precomputed, stored in lookup tables and reused in each cell during calculation of terms $\left\langle q,\colvec{2}{c_i^1}{c_j^2} \right\rangle$ and $\left\langle q,\colvec{3}{r_1}{\vdots}{r_M} \right\rangle$. Given that dot-products are precomputed calculation of these terms can be done in $O(M)$ operations.

Also note that term $\left\|\colvec{2}{c_i^1}{c_j^2}+\colvec{3}{r_1}{\vdots}{r_M}\right\|^2$ is query-independent. It can be precomputed before search and henceforth used via lookup tables. Due to nice properties of PQ this term can be further simplified:

\begin{gather}
\label{eq:decomposition}
\left\|\colvec{2}{c_i^1}{c_j^2} + \colvec{3}{r_1}{\vdots}{r_M}\right\|^2 = \left\|c_i^1 + \colvec{3}{r_1}{\vdots}{r_{\frac{M}{2}}}\right\|^2 + \left\|c_j^2 + \colvec{3}{r_{\frac{M}{2}+1}}{\vdots}{r_M}\right\|^2 = \\
\left\|c_i^1\right\|^2 + \left\|c_j^2\right\|^2 + \sum\limits_{k=1}^{M}\left\|r_k\right\|^2 +
2\sum\limits_{k=1}^{\frac{M}{2}}\langle c_i^1, r_k\rangle + 
2\sum\limits_{k=\frac{M}{2}+1}^{M}\langle c_j^2, r_k\rangle \nonumber
\end{gather}

Thus, it is enough to store squared norms of codebook centroids and dot-product of indexing and compression centroids in lookup tables. Given these values calculation of these term can be also perfromed in $O(M)$ operations. As a result, all terms in distance evaluation expression~\eq{distance} can be calculated in $O(M)$ operations, which is substantially faster than in Multi-D-ADC, especially for high-dimensional data. Below we prove this claim experimentally.

Overall, for a given database the FBPQ and the Multi-D-ADC will have the same index structure and will visit the same points in the same order during search. Hence the FBPQ will always provide the same recall as the Multi-D-ADC, however, this now can be done several times faster.
\\
\\

\subsection{HBPQ}

\begin{table*}
\small
\centering
\addtolength{\tabcolsep}{-1pt}
\renewcommand\arraystretch{1.3}
\begin{tabular}{|c|c|ccc|c|c|}
\hline
System& $l$ &  R@1 & R@10 & R@100 & Time(ms)& Memory(Gb)\\
\hline
\multicolumn{7}{|c|}{\bf BIGANN, 1 billion SIFTs,  8 bytes per vector}\\
\hline
OMulti-D-OADC & 10000 & 0.179 & 0.523 & 0.757 & 4.9 & 13\\
Hierarchical-BPQ & 10000 & {\bf 0.268} & {\bf 0.644} & {\bf 0.776} & 6.2 & 15\\
\hline
OMulti-D-OADC & 30000 & 0.184 & 0.549 & 0.853 & 13.8 & 13\\
Hierarchical-BPQ & 30000 & {\bf 0.280} &{\bf 0.704} & {\bf 0.894} & 16.1& 15\\
\hline
OMulti-D-OADC & 100000 & 0.186 & 0.556 & 0.894 & 41.3 & 13\\
Hierarchical-BPQ & 100000 & {\bf 0.286} & {\bf 0.729} & {\bf 0.952} & 49.6 & 15\\
\hline
\multicolumn{7}{|c|}{\bf BIGANN, 1 billion SIFTs,  16 bytes per vector}\\
\hline
OMulti-D-OADC & 10000 & $0.342_{(0.345)}$ & $0.714_{(0.725)}$ & $0.781_{(0.794)}$ & $ 5.6_{(6.9)}$ & 21\\
Hierarchical-BPQ&10000& {\bf 0.421}&{\bf 0.755}&{\bf 0.782}& 6.8& 23\\
\hline
OMulti-D-OADC&30000& $0.360_{(0.366)}$&$0.797_{(0.807)}$&$0.905_{(0.913)}$& $14.9_{(16.9)}$& 21\\
Hierarchical-BPQ&30000& {\bf 0.454}&{\bf 0.862}&{\bf 0.908}& 18.7& 23\\
\hline
OMulti-D-OADC&100000& $0.368_{(0.373)}$&$0.835_{(0.841)}$&$0.972_{(0.973)}$& $49.5_{(51.5)}$& 21\\
Hierarchical-BPQ&100000& {\bf 0.467}&{\bf 0.914}&{\bf 0.976}& 66.2& 23\\
\hline
\end{tabular}
\caption{Comparison of the bilayer systems accuracy: OMulti-D-OADC and the optimized HBPQ for SIFT1B dataset. Numbers in brackets were reported in \cite{OpqTr}, we see that out re-implementation is quite comparable to \cite{OpqTr}. $l$ is a number of candidates reranked by both systems.}
\label{tab:HBPQ}
\end{table*}
\addtolength{\tabcolsep}{-1pt}

The second modification, the HBPQ, provides substantially higher recall w.r.t.\ the Multi-D-ADC at a cost of a slight increase in runtime. The HBPQ is motivated by the fact that the use of global codebooks for displacements encoding in the Multi-D-ADC (and OMulti-D-OADC) leads to large approximation errors, since the displacements in different cells are distributed differently. Thus, first row of \fig{global_vs_local} shows distributions of first two principal components for displacements in a few cells of the second-order multi-index with $T = 2^{14}$ built for SIFT1B dataset. The different nature of distributions corresponding to different cells can be clearly seen, and it thus suggests that using \emph{different} codebooks to encode displacements within different cells should improve the accuracy of the encoding.

The second motivating observation discussed below in more detail is that the local codebooks in the HBPQ can still be shared across subsets of cells in a natural way, which results in a reasonable memory consumption, and makes the whole system attractive from the viewpoint of the memory-search accuracy tradeoff.

Below we introduce and discuss the details of the HBPQ system. As before, we describe the HBPQ as an extension of the second-order Multi-D-ADC system but the generalization to other orders is straightforward. We will continue with the notation from \sect{multidadc}.

The main difference of the HBPQ and the Multi-D-ADC is a process of codebooks learning. Indexing-level codebooks $C^1$ and $C^2$ and quantizers $q^1$  and $q^2$ are obtained in the same way via $K$-means clustering of first and second halves of learn points respectively. 

The quantizer $q^1$ defines a partition of $\frac{D}{2}$-dimensional space corresponding to the first half of dimensions into $T$ Voronoi cells $U_1, \ldots, U_T$:

\begin{gather}
\label{eq:Ui}
U_i = \{ x \in \mathbf{R}^{\frac{D}{2}} | \; q^1(x) = c_i^1 \}
\end{gather}

For the $i$-th Voronoi cell $U_i$, the local PQ codebooks is learned on the set of displacements of the vectors falling into this cell w.r.t.\ the corresponding cell centroid:
\begin{gather}
\label{eq:pi1}
\{p_i^1 - c_i^1 \; | \; q^1(p_i^1) = c_i^1, \; [p_i^1\;p_i^2] \in L\}
\end{gather}

If total number of PQ-codebooks is $M$, then a number of codebooks coding halves of vectors should be $\frac{M}{2}$. Thus, for each cell $U_i$ the HBPQ learns local codebooks $(Q_1^i, \dots, Q_{M/2}^i)$.
The local codebooks in each cell $V_j$ that corresponds to the second halves of dimensions $\{ x \in \mathbf{R}^{\frac{D}{2}} | \; q^2(x) = c_j^1 \}$ are obtained in a similar way and are denoted as $(S_1^j, \dots, S_{M/2}^j)$.


Overall, the set of HBPQ codebooks consists of

\begin{enumerate}
\item Two indexing-level codebooks $C^1$ and $C^2$, \\ $|C^1| = |C^2| = T$
\item $TM$ compression-level codebooks $\{Q_1^i, \dots, Q_{M/2}^i\}_{i=1}^T$ and $\{S_1^i, \dots, S_{M/2}^i\}_{i=1}^T$
\end{enumerate}

The index construction process in the HBPQ is the same as in the Multi-D-ADC except the fact that displacements in a cell $W_{ij}$ are encoded via codebooks $(Q_1^i, \dots, Q_{M/2}^i)$ and $(S_1^j, \dots, S_{M/2}^j)$.

For each point $p$ falling into the cell $W_{ij}$, the HBPQ encodes the $D$-dimensional vector of displacement:
\begin{gather}
\label{eq:residual}
d = p - \left[c_i^1 \; c_j^2\right] = \left[p^1 - c_i^1 \; p^2 - c_j^2\right] = \left[d^1 \; d^2\right]
\end{gather}

The PQ encoding of a displacement $d$ is a concatenation of encodings of $\frac{D}{2}$-dimensional displacements $d^1$ and $d^2$. The HBPQ uses the local codebooks $(Q_1^i, \dots, Q_{M/2}^i)$ to encode $d^1$ and the local codebook $(S_1^j, \dots, S_{M/2}^j)$ to encode $d^2$. This encoding is expected to be accurate because $(Q_1^i, \dots, Q_{M/2}^i)$ and $(S_1^j, \dots, S_{M/2}^j)$ are learnt locally and are therefore adapted to distributions of displacements within the corresponding cells.

From the geometric viewpoint, the local codebooks subdivide each Voronoi cell into finer cells (which are in turn products of Voronoi cells corresponding to PQ components). Finally, the ultimate space subdivision is defined by cartesian products of those finer cells corresponding to different halves of the dimensions.

Query processing in the HBPQ is performed analogously to the Multi-D-ADC except that displacements are reconstructed more precisely via local codebooks. The complexity of distance evaluation in the HBPQ is also $O(D)$ operations.
The use of the local codebooks allows HBPQ to perform more accurate reconstructions, and to outperform Multi-D-ADC (e.g.\ return the entry corresponding to the true nearest neighbor within the top candidates more often) due to the more reliable reranking.

\subsubsection{Optimized HBPQ}

The product quantization at both levels of the HBPQ can be replaced with the optimized product quantization (OPQ). For this, one orthogonal transformation is applied to original vectors before the first PQ step. At the fine level, two separate transformations are applied to $\frac{D}{2}$-dimensional displacements $r^1$ and $r^2$ before encoding. Matrices of these transformations are learnt jointly with the codebooks (for more details see \cite{OpqTr}).
\\
\\

\subsection{Additional memory consumption analysis}

Both FBPQ and HBPQ require additional memory to keep lookup tables and local codebooks respectively. Now we quantify these memory costs for both systems.

The second-order FBPQ system with $|C^1| = |C^2| = T$ and $|R_1| = \dots = |R_M| = K$ requires two tables keeping dot-products $\langle c_i^1, r_k\rangle$ and $\langle c_j^2, r_k\rangle$ with a total size of $TMK$ floating-point numbers. Precomputed norms of codewords require  $(2T + MK)$ floating-point numbers. For a typical second-order system \cite{Babenko12, OpqTr} having two codebooks of size $2^{14}$ for indexing and $8$ codebooks of size $256$ for compression FBPQ requires only $128$ Mb for dot-products and less than $1$ Mb for norms.

Now we consider the issue of memory requirements for storing the local codebooks in HBPQ. Generally speaking, local codebooks are a bottleneck on efficiency. If one would want to maintain local codebooks within most of existing systems, it would require the amount of memory which is linear in number of cells ($T^2$ for the Multi-D-ADC). In the case of the HBPQ, however, this amount of memory is reduced by a factor of $T$  and overall the amount of memory spent on local codebooks grows as $T$. This is achieved because the local codebooks are effectively shared across groups of cells $W_{ij}$ that have the same $i$ or $j$-index (so, in some sense, these codebooks are \emph{semi-local}). We show in the experiments that for typical values of $T$ suggested in previous works, the increase of memory usage is small and the total memory consumption is comparable with the systems in \cite{Babenko12, OpqTr}. We give some examples in the \sect{hbpq_experiments}.

\section{Experiments}

\begin{table*}
\centering
\addtolength{\tabcolsep}{1pt}
\renewcommand\arraystretch{1.5}
\begin{tabular}{|c|ccc|c|c|}
\hline
Method& R@1 & R@10 & R@100 & Time& Memory\\
\hline
IVFADC+R\cite{Jegou11b} &0.262&0.701&0.962& 116 ms& 20 Gb\\
\hline
Multi-D-ADC\cite{Babenko12} &0.334&0.793&0.959& 49 ms& 21 Gb\\
\hline
OMulti-D-OADC\cite{OpqTr} &0.373&0.841&0.973& 51.5 ms& 21 Gb\\
\hline
 Joint inverted indexing\cite{Joint13}& --- & --- &0.938& 11.8 ms& 80 Gb\\
\hline
HBPQ &{\bf 0.467}&{\bf 0.914}&{\bf 0.976}& 66.2 ms& 23 Gb\\ 
\hline
\end{tabular}
\caption{Reported performance of existing methods on SIFT1B (BIGANN) dataset containing a billion of 128-dimensional SIFT descriptors. All methods use 16 bytes per vector. The optimized HBPQ system with local codebooks establishes new state-of-the-art performance on SIFT1B.}
\label{tab:comparison}
\end{table*}
\addtolength{\tabcolsep}{-2pt}

\begin{figure}
\centering
\noindent\includegraphics[width=8.5cm]{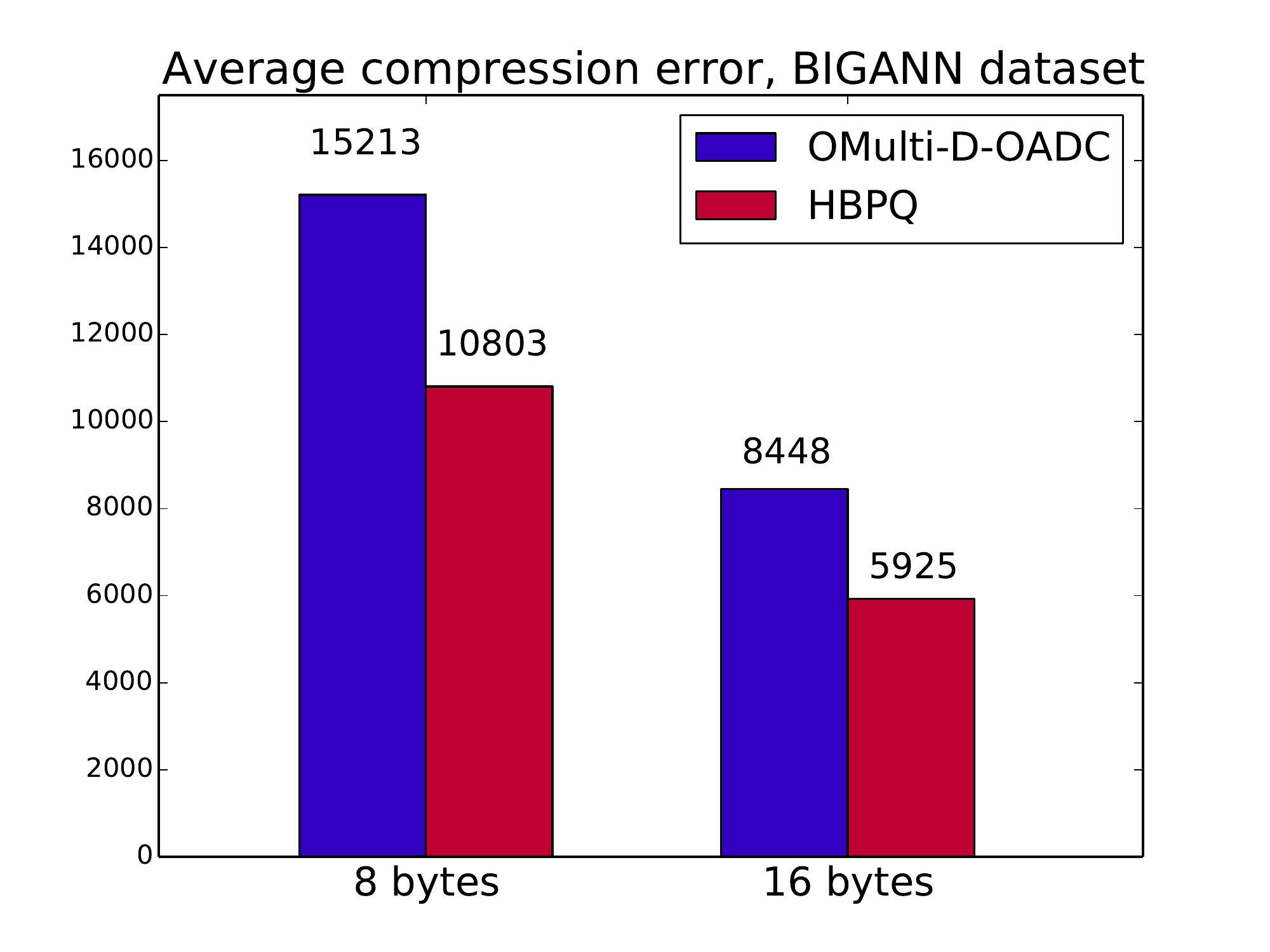}
\caption{Average encoding errors of SIFT1B dataset for the OMulti-D-OADC system with global codebooks and the optimized HBPQ with local codebooks. For both typical code lengths (8 and 16 bytes), the errors associated with local codebooks are considerably (30\%) lower than with global codebooks.}
\label{fig:errors}
\end{figure}

We perform the bulk of experiments on the BIGANN dataset~\cite{Jegou11a} containing one billion 128-dimensional SIFT vectors in the base set and 100 millions vectors in the learning set (the dataset is also known as SIFT1B). The dataset also comes with 10,000 queries with known true Euclidean nearest neighbors among the base set. All codebooks for multi-index and product quantization were learnt on the provided hold-out learning set. We also perform additional experiments on a more higher-dimensional GIST50M dataset which is a subset of 80M Tiny images \cite{Torralba08b}. In our experiments, this dataset contains $50$ millions of $384$-dimensional GIST descriptors \cite{Oliva01} in a base set, 1000 queries and $29$ millions of GIST descriptors in a hold-out learn set.

In all experiments we compare the previous state-of-the-art system OMulti-D-OADC and the optimized FBPQ/HBPQ (also based on optimized product quantization) while fixing the main parameters to their suggested values in the previous works (the number of cells, the number of PQ bytes, etc.). In particular, we use $T=2^{14}$ (hence, number of cells is $2^{28}$) for SIFT1B and $T=2^{10}$ (hence, number of cells is $K=2^{20}$) cells for GIST50M. We use our re-implementation of system \cite{OpqTr} and experiments show that is is highly comparable with the origin. We also follow the previous works \cite{Jegou11b, Babenko12, OpqTr} and use the \emph{Recall@T} as a main measure of the NN-search system quality. The Recall@T is defined as a probability of finding the true nearest neighbor in a list of length $T$ returned by the system.

\begin{figure}
\centering
\hspace{-1mm}\noindent\includegraphics[width=8.5cm]{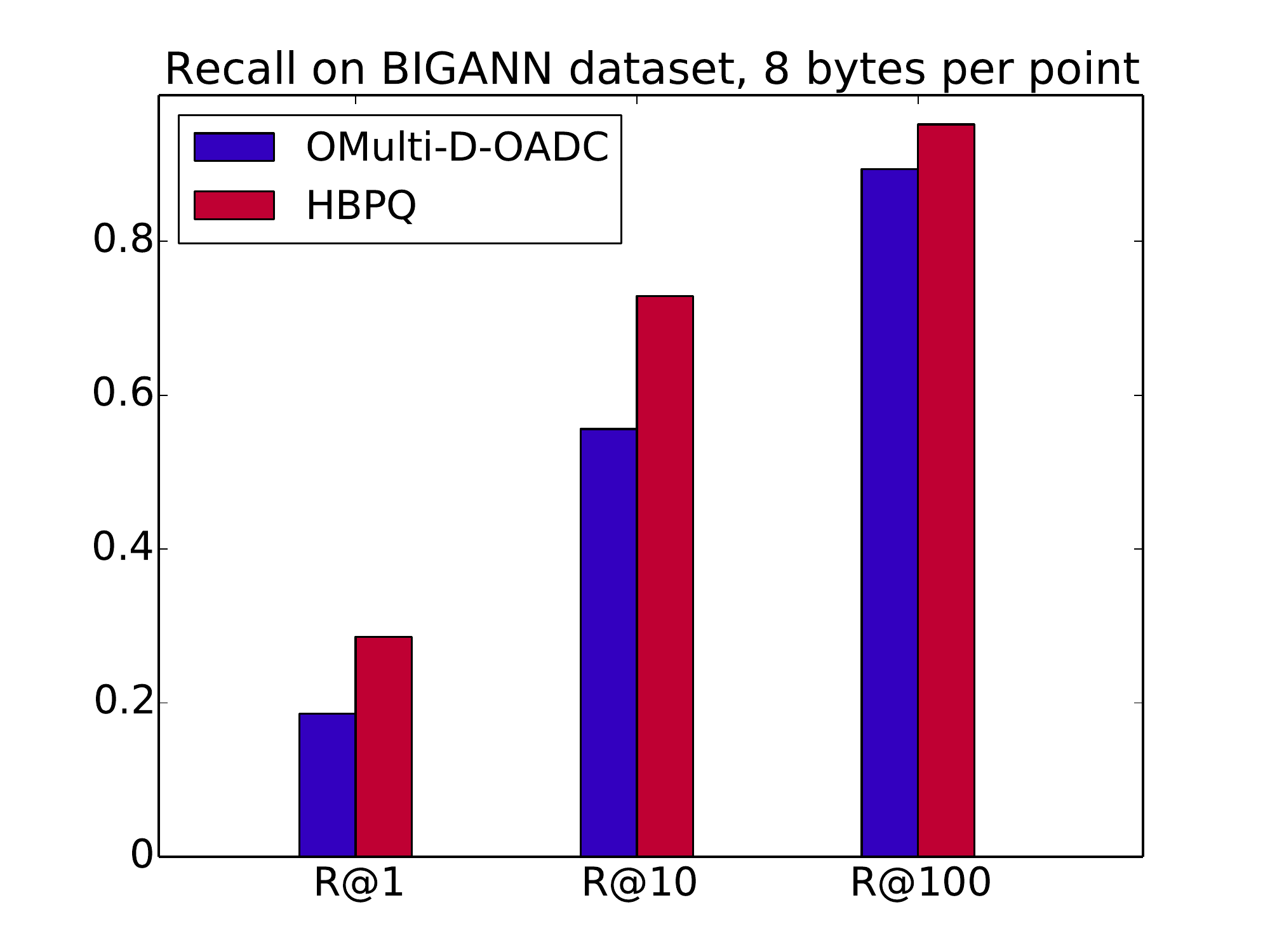}
\caption{The recall@T for the nearest neighbor search for the SIFT1B dataset for the OMulti-D-OADC and the optimized HBPQ systems. The local displacements are encoded with 8 bytes. The use of local codebooks allows to achieve the 10\% gain in Recall@1 and the 17\% gain in Recall@10. This gain in recall is due to better encoding via local codebooks.\vspace{-4mm}}
\label{fig:recall8}
\end{figure}

\subsection{Improved runtime with FBPQ}

In this section we show that accuracy of the OMulti-D-OADC can be achieved several times faster with the optimized FBPQ. \tab{FBPQ} shows the performance of both systems for different short-list lengths. As expected, systems provide the same recall values.

We can see that runtime of the FBPQ is several times smaller for both datasets. This gain is more impressive for GIST50M as GIST descriptors are $3$ times longer than SIFT.
 
Also note that additional memory consumption of the FBPQ is negligible for both datasets.
\\

\subsection{Improved accuracy with HBPQ}

Now we give experimental comparison of the OMulti-D-OADC and the optimized HBPQ systems.

\subsubsection{Encoding error}

We start by comparing the encoding errors across two systems. In \fig{errors}, we plot the average encoding error for SIFT1B dataset with the typical code lengths (8 and 16 bytes). \fig{errors} demonstrates that the usage of HBPQ with local codebooks reduces the approximation error by 30\% for both code lengths. Below we show that this gain in encoding accuracy results in far better performance of the nearest neighbor search.

\subsubsection{Nearest neighbor search}
\label{sect:hbpq_experiments}

We now measure how the improvement in the encoding accuracy affects the accuracy of the nearest neighbor search.

We once again compare the performance of the second-order OMulti-D-OADC system with $2^{28}$ cells as in \cite{OpqTr} and the HBPQ with the same $T$. \tab{HBPQ} demonstrates that the HBPQ outperforms OMulti-D-OADC with global codebooks considerably. Recall@1 increased by $10\%$ for both code lengths and Recall@10 increased by $8-17\%$. This gain comes at a price of $2$Gb of additional memory for local codebooks in HBPQ (which is $15\%$ addition for ADC-$8$ and $9\%$ addition for ADC-$16$). In many scenarios, this increase in memory requirements would not be substantial and would not influence the applicability of the HBPQ. We also observed a small ($20\%$) increase in runtime that is probably due the fact that local codebooks are too large to fit into the CPU cache unlike global codebooks.

Results for GIST50M dataset are shown in a \tab{gist_performance}. In this case, the gain in recall is less than for SIFT1B due to smaller $T$ resulting in decreased specificity of local codebooks.

{\bf Comparison to the prior art.} \tab{comparison} compares all reported results for the SIFT1B dataset in terms of recall, runtime and memory consumption. Generally, the HBPQ offers a favourable tradeoff in terms of search accuracy, memory consumption, and the query time, and arguable establishes the new state-of-the-art for this dataset. 

\begin{table}
\small
\centering
\addtolength{\tabcolsep}{-0.3pt}
\renewcommand\arraystretch{1.5}
\begin{tabular}{|c|c|ccc|c|c|}
\hline
System& $l$ &  R@1 & R@10 & R@100 & Time& Memory\\
\hline
\multicolumn{7}{|c|}{\bf GIST50M, 50 millions GISTs,  8 bytes per vector}\\
\hline
OMulti-D-OADC & 10000 & 0.317 & 0.454 & 0.569 & 6.3 ms & 0.57 Gb\\
HBPQ & 10000& {\bf 0.321} & {\bf 0.489} & {\bf 0.581} & 7.3 ms & 0.95 Gb\\
\hline
OMulti-D-OADC & 30000& 0.323 & 0.496 & 0.659 & 18.5 ms & 0.57 Gb\\
HBPQ & 30000 & {\bf 0.336} & {\bf 0.536} & {\bf 0.691} & 21.9 ms & 0.95 Gb\\
\hline
OMulti-D-OADC & 100000& 0.327 & 0.512 & 0.711 & 61.4 ms & 0.57 Gb\\
HBPQ & 100000 & {\bf 0.347} & {\bf 0.556} & {\bf 0.773} & 69.7 ms & 0.95 Gb\\
\hline
\end{tabular}
\caption{Comparison of the bilayer systems accuracy: OMulti-D-OADC and HBPQ for GIST50M dataset.  $l$ is a number of candidates reranked by both systems.}
\label{tab:gist_performance}
\end{table}
\addtolength{\tabcolsep}{-1pt}

\section{Conclusion}

We have proposed and evaluated two systems for efficent approximate nearest neighbor search. The first system, FBPQ, performs search several times faster than the top-performing Multi-D-ADC system and provides new state-of-the-art on BIGANN dataset in terms of runtime. The FBPQ can be used in any application which includes the Multi-D-ADC without any additional costs. The second system (HBPQ) substantially outperforms existing systems in terms of recall. The improvement over Multi-D-ADC comes at the price of a small increase in memory consumption, which should be tolerable for most applications.

One possible limitation of the HBPQ is the fact that it requires a lot of data for learning --- we used full avaliable learn set of SIFT1B (100 millions of vectors) to learn codebooks for the second order multi-index with $T=2^{14}$. This is however a limitation only if one insists on training all codebooks on a separate set (to avoid overfitting). In some scenarios, it may be in fact beneficial to fit the ``test'' data over which the search will be performed, and the HBPQ provides more flexibility to do that compared to systems using global codebooks. Generally, the larger is the scale of the dataset, the more natural becomes the use of local codebooks and the HBPQ system.

\bibliographystyle{abbrv}
\bibliography{acmmm14}
\end{document}